\documentclass[runningheads]{llncs}
\usepackage[T1]{fontenc}
\usepackage{graphicx,verbatim}
\usepackage{amsmath}
\usepackage{bm}
\usepackage{amssymb}
\usepackage{esvect}
\usepackage{mathtools}
\usepackage{booktabs}
\usepackage{comment}
\usepackage{placeins}
\usepackage{xcolor}
\usepackage[caption=false]{subfig}
\usepackage{array}
\usepackage{makecell} 
\usepackage{siunitx}
\usepackage{threeparttable}
\usepackage{adjustbox}
\usepackage{xfp}
\usepackage{xcolor}
\newcommand{\cellsize}{\footnotesize} 

\newcommand{\pmNum}[2]{\ensuremath{\num{#1}\,\pm\,\num{#2}}}

\newcommand{\numSSIM}[1]{\num[round-mode=places,round-precision=1]{\fpeval{100*(#1)}}}
\newcommand{\numSSIMstd}[1]{\num[round-mode=places,round-precision=1]{\fpeval{100*(#1)}}}
\newcommand{\numPSNR}[1]{\num[round-mode=places,round-precision=2]{#1}}
\newcommand{\numPSNRstd}[1]{\num[round-mode=places,round-precision=2]{#1}}

\newcommand{\pmSSIM}[2]{\ensuremath{\numSSIM{#1}\pm\numSSIMstd{#2}}}
\newcommand{\pmPSNR}[2]{\ensuremath{\numPSNR{#1}\pm\numPSNRstd{#2}}}

\newcommand{\spcell}[4]{%
  \makecell[c]{\cellsize \pmSSIM{#1}{#2}\\ \cellsize \pmPSNR{#3}{#4}}%
}

\newcommand{\spcellbest}[4]{%
  \makecell[c]{%
    \cellsize \textbf{\numSSIM{#1}}\,\ensuremath{\pm}\,\numSSIMstd{#2}\\
    \cellsize \textbf{\numPSNR{#3}}\,\ensuremath{\pm}\,\numPSNRstd{#4}%
  }%
}

\newcommand{\spcellsecond}[4]{%
  \makecell[c]{%
    \cellsize \numSSIM{#1}\textsuperscript{\dag}\,\ensuremath{\pm}\,\numSSIMstd{#2}\\
    \cellsize \numPSNR{#3}\textsuperscript{\dag}\,\ensuremath{\pm}\,\numPSNRstd{#4}%
  }%
}

\begin{document}
\title{Heterogeneity-Adaptive Diffusion Schrödinger Bridge for PET-Guided Whole-Body MRI Translation}
\titlerunning{Heterogeneity-Adaptive Schr\"odinger Bridge}

%

\author{
Chengbo Wang\inst{1}\textsuperscript{\(\dagger\)} \and
Jiacheng Yu\inst{1}\textsuperscript{\(\dagger\)} \and
Linjie Bian\inst{2}\textsuperscript{\(\dagger\)} \and
Ming Qi\inst{2} \and
Xiaosheng Liu\inst{2} \and
Tongtong Che\inst{3} \and
Jichang Zhang\inst{3} \and
Shuyu Li\inst{3} \and
Shaoli Song\inst{2}\textsuperscript{*}and
Xiuying Wang\inst{1}\textsuperscript{*}
}

\authorrunning{C. Wang et al.}

\institute{
The University of Sydney, Sydney, Australia
\and
Department of Nuclear Medicine, Fudan University Shanghai Cancer Center, Shanghai, China
\and
State Key Laboratory of Cognitive Neuroscience and Learning, Beijing Normal University, Beijing, China
}

\maketitle

\begingroup
\renewcommand{\thefootnote}{}
\footnotetext{
\textsuperscript{\(\dagger\)} contributed equally to this work. \texttt{cw643@cantab.ac.uk}, \texttt{mryu888@foxmail.com},
\texttt{blijer@163.com}
}
\footnotetext{
\textsuperscript{*} corresponding authors.}.
\endgroup

\begin{abstract}


While whole-body multimodal medical imaging scanners have
been increasingly recognized for more effective medical applications, the
excessive long acquisition time in PET–MR scanning is a major obstacle
in more efficient clinical practice. Deep learning–based MRI translation
provides a potential solution to reduce scan duration. However, current
models often focus on specific anatomical regions and face challenges
for whole-body scans that consists of highly heterogeneous feature distributions mainly due to (1) 
different anatomical regions across whole-body, and (2) lesions or pathological tissues. This paper tackles the challenges through a novel Heterogeneity-Adaptive Diffusion Schrödinger Bridge (HA-DSB) framework. By explicitly modeling translation as stochastic transport between source and target distributions, HA-DSB incorporates region context embeddings derived from a vision-language model (VLM) to enable region-specific modeling. To enhance fidelity of the pathological tissue, lesion-aware metabolic prior from PET is integrated directly into the bridge dynamics through a dual-stage guidance mechanism. Specifically, a PET-guided noise modulation module adaptively scales spatial diffusion perturbations during the forward process, while PET features are leveraged during the reverse process to selectively amplify lesion-relevant structures via an attention mechanism. Experiments demonstrate the superiority of our method across different body regions in whole-body MRI translation and show improved translation quality in lesion areas under PET guidance. Our code is available at {\color{blue}\url{https://github.com/xyw-medical-research/HADSB}}.


\keywords{Medical Image Translation  \and PET/MR Imaging \and Diffusion Schrödinger Bridge \and Whole-body MR Translation}

\end{abstract}
\section{Introduction}
\begin{figure}[t]
    \centering
    \includegraphics[width=1\linewidth]{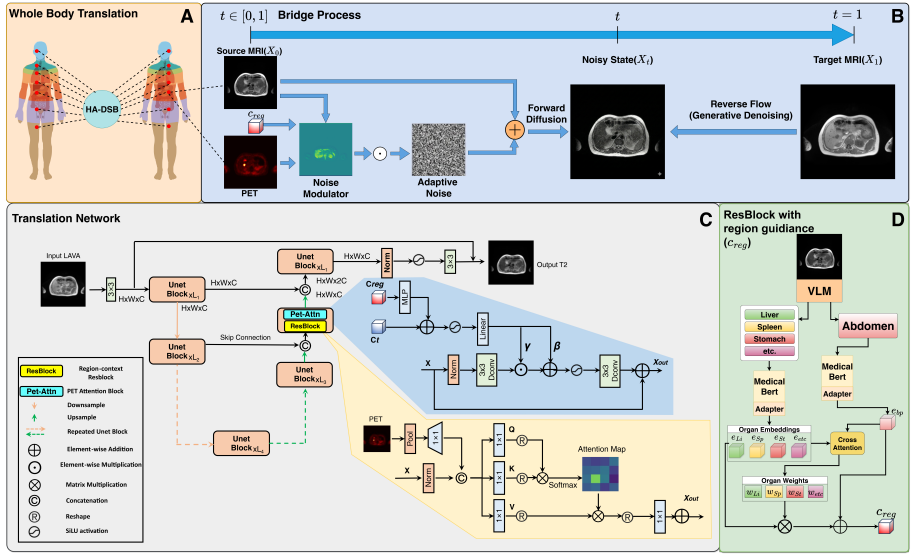}
    \caption{Overview of HA-DSB. (a) Whole-body MR translation setting. (b) Diffusion Schr\"odinger bridge from LAVA to T2 with noise modulation jointly conditioned on PET and region context $\mathbf{c}_{\text{reg}}$. (c) UNet translation backbone with region-conditioned residual blocks and multi-scale PET attention. (d) Construction of the region context embedding $\mathbf{c}_{\text{reg}}$.}
    \label{fig:method}
\end{figure}

Hybrid positron emission tomography combined with magnetic resonance imaging (PET/MR) is an emerging imaging modality that enables simultaneous acquisition of metabolic and high-resolution anatomical information, producing intrinsically co-registered PET and multi-sequence MR images while reducing radiation exposure compared with PET/CT \cite{Mass}. Whole-body PET/MR supports comprehensive multi-bed imaging from the head and neck to the mid-thigh and has become increasingly important in clinical applications such as cancer metastasis evaluation \cite{Catalano}. However, whole-body PET/MR scan time typically ranges from 60–90 minutes, largely due to multiple MR sequences, which limits clinical efficiency and increases patient discomfort. Deep learning–based MR translation offers a practical solution by synthesizing time-consuming modalities from fast-acquired sequences, and recent diffusion-based approaches have demonstrated promising performance in MR image translation ~\cite{Ozbey,XingZ,Dalmaz_2022,arslan2024selfconsistentrecursivediffusionbridge}.\\
\indent  However, most MR translation focus on specific anatomical regions such as the brain in public datasets \cite{Jiang}, and extending these models to whole-body MR translation in integrated PET/MR settings remains challenging. Whole-body scans exhibit strong cross-region heterogeneity, with substantial variations in tissue composition and signal characteristics across anatomical regions~\cite{Dzyu}. Such non-stationary and region-dependent intensity distributions significantly increase the difficulty of learning a unified modality mapping~\cite{ZhengJ,ZhengJia}. For example, GANs are known to suffer from mode collapse in multi-modal distributions~\cite{GongY,LuoY}, while diffusion models may be biased toward dominant feature modes~\cite{Um}. Without explicit region conditioning, these models may struggle to capture region-specific structures in the whole body MR translation.\\
\indent Another critical yet underexplored challenge in MR translation is the accurate reconstruction of pathological regions~\cite{ZhangX,Daya}, which are often of primary clinical importance. Compared to normal tissues, lesions exhibit highly heterogeneous appearances and complex cross-modality signal relationships. Most existing models do not explicitly differentiate pathological changes from normal tissues, causing lesion signals to be biased toward the dominant healthy distribution and leading to degraded pathological fidelity. \\
\indent In this work, we propose a Heterogeneity-Adaptive Diffusion Schrödinger Bridge (HA-DSB) with PET guidance for whole-body MR translation (Fig.~\ref{fig:method}). Unlike conventional diffusion-based translation methods that rely on implicit noise prediction, HA-DSB explicitly optimizes a stochastic transport plan between source and target MR distributions \cite{Ozbey,Gung}. This explicit transport formulation is better suited for whole-body MR translation, where strong cross-region domain shifts make implicit alignment prone to over-smoothing and modality ambiguity. To address the pronounced region-dependent distribution shifts in whole-body MR translation, we further introduce region context embeddings derived from structured descriptions generated by the large vision-language model and fuse them with diffusion time embeddings, enabling fine-grained, region-specific bridge modeling and consistently improving translation performance across anatomical regions. Furthermore, to improve pathological fidelity, we exploit PET as a lesion-aware prior, which is naturally available in integrated PET/MR due to intrinsic co-registration. Since lesions typically manifest as focal high-uptake signals in PET, we introduce a PET-guided noise modulation mechanism that applies spatially adaptive corruption in the forward process. This spatially adaptive perturbation redistributes the stochastic transport difficulty, encouraging the model to learn pathology-sensitive mappings rather than being biased toward dominant healthy structures. A multi-scale PET-guided attention module is further incorporated into the reverse process to amplify lesion-relevant features, improving translation fidelity in pathological areas. Our main contributions include:
\begin{itemize}
\item We propose a Heterogeneity-Adaptive Diffusion Schrödinger Bridge (HA-DSB) that models whole-body MR translation as an explicit distribution transport problem, achieving more accurate and stable cross-modality mapping across anatomically diverse regions compared to conventional noise-prediction-based diffusion models.
\item We introduce region context embeddings generated by a large vision-language model and fuse them with diffusion time embeddings as conditioning signals, enabling the bridge to adapt its region-specific transport dynamics.
\item We design a PET-guided lesion enhancement mechanism in which the forward corruption process is spatially modulated by PET uptake and region context, while multi-scale PET-aware attention improves lesion-relevant feature recovery during reverse denoising.
\end{itemize}

\section{Methodology}
\subsection{Diffusion Schr\"odinger Bridge.}

Diffusion Schr\"odinger bridge (DSB) \cite{ChenT,LiuG} constructs a stochastic process that transports samples between two endpoint distributions. In image-to-image translation, the bridge provides an endpoint-to-endpoint alternative to the usual Gaussian-prior diffusion models. The forward and backward process can be described as:

\begin{equation}
\small
\mathrm{d}X_t =
\left[f + g^2 \nabla_x \log \Psi(t, X_t)\right]\mathrm{d}t
+ g \, \mathrm{d}W_t, \quad
\mathrm{d}X_t =
\left[f - g^2 \nabla_x \log \hat{\Psi}(t, X_t)\right]\mathrm{d}t
+ g \, \mathrm{d}W_t.
\end{equation}
where $X_0$ and $X_1$ denote the source and endpoint states, and $\Psi$ and $\hat{\Psi}$ are time-dependent Schr\"odinger potentials \cite{ChenT}. We adopt the I$^2$SB instantiation proposed in \cite{LiuG}, assuming paired samples $(X_0, X_1)$ drawn from the joint distribution $p(X_0, X_1)=p_A(X_0)p_B(X_1\mid X_0)$ and setting the drift term $f=0$. Under this formulation, the Schrödinger Bridge can be efficiently trained and inferred within the score-based generative modeling (SGM) framework. The analytic posterior of DSB given a boundary pair $(X_0, X_1)$ can be expressed as $ q(X_t \mid X_0, X_1) = \mathcal{N}\!\big(X_t;\, \mu_t(X_0, X_1),\, \Sigma_t\big) $.
\begin{equation}
\mu_t
= \frac{\bar{\sigma}_t^{2}}{\bar{\sigma}_t^{2} + \sigma_t^{2}}\, X_0
+ \frac{\sigma_t^{2}}{\bar{\sigma}_t^{2} + \sigma_t^{2}}\, X_1,
\qquad
\Sigma_t
= \frac{\sigma_t^{2}\,\bar{\sigma}_t^{2}}{\sigma_t^{2} + \bar{\sigma}_t^{2}}\, \mathbf{I}
\label{eq:dsb_posterior}
\end{equation}
where $\sigma_t^{2} \coloneqq \int_{0}^{t} \beta_{\tau}\,\mathrm{d}\tau$ and 
$\bar{\sigma}_t^{2} \coloneqq \int_{t}^{1} \beta_{\tau}\,\mathrm{d}\tau$
denote the variances accumulated on the two sides of time $t$. Unlike conventional image-to-image Schr\"odinger Bridge models, our method adopts a heterogeneity-adaptive conditional formulation that integrates region context embedding and PET-derived cues. The final training objective is defined as follows, where $c$ represents the conditioning information.

\begin{equation}
\mathcal L(\theta)=
\mathbb E\Big[\big\|\varepsilon_\theta(X_t,c,t)-\tfrac{X_t-X_0}{\sigma_t}\big\|_2^2\Big].
\end{equation}

\subsection{Region context embedding guidance.}
To address anatomical heterogeneity in whole-body translation, we introduce a region context embedding $\mathbf{c}_{\text{reg}} \in \mathbb{R}^d$ that adapts the bridge dynamics to region-dependent structural variability. For each slice, we assign a body-location label $l_{\text{bp}}$ that partitions the body into $K=11$ regions. A frozen PubMedBERT~\cite{gu2021pubmedbert} encodes these labels into semantic features, which are linearly projected and combined with learnable embeddings to obtain the body-location representation $\mathbf{e}_{\text{bp}} \in \mathbb{R}^d$.

To capture intra-region variability, each slice is additionally associated with a set of organ labels, which are encoded by the same frozen encoder to produce organ embeddings $\{\mathbf{e}_{\text{org}}^i\}_{i=1}^{M}$, where $\mathbf{e}_{\text{org}}^i \in \mathbb{R}^d$ and $M$ is the number of organs present in the slice. We fuse body-location and organ information via multi-head cross-attention, where $\mathbf{e}_{\text{bp}}$ serves as the query and organ embeddings act as the keys and values. The organ information is then added to the body-location representation as a residual correction, forming the final region context embedding $\mathbf{c}_{\text{reg}}$:
\begin{equation}
\small
\alpha_i
=
\mathrm{Softmax}
\left(
\frac{
(\mathbf{W}_Q \mathbf{e}_{\text{bp}})^\top
(\mathbf{W}_K \mathbf{e}_{\text{org}}^i)
}{
\sqrt{d_h}
}
\right),
\qquad
\mathbf{e}_{\text{attn}} = \sum_{i=1}^{M} \alpha_i \, \mathbf{W}_V \mathbf{e}_{\text{org}}^{i},
\end{equation}
\begin{equation}
\small
\label{eq:residual_connection}
\mathbf{c}_{\text{reg}} = \mathbf{e}_{\text{bp}} + \mathrm{MLP}\!\left(\mathrm{LayerNorm}(\mathbf{e}_{\text{attn}})\right),
\end{equation}
where $\mathbf{W}_Q,\mathbf{W}_K,\mathbf{W}_V \in \mathbb{R}^{d_h \times d}$ are learnable projections, with $d_h=d/H$ and $H=4$. The resulting $\mathbf{c}_{\text{reg}}$ encodes both coarse body-location and fine-grained organ information. It is added to the diffusion timestep embedding $\mathbf{e}_t \in \mathbb{R}^d$ and passed through an MLP to generate the scale and shift parameters in each residual block.

To avoid manual annotation, we use Google Gemini 3 Pro to generate body-location and organ labels. The reliability of body-location labels was empirically verified, achieving approximately 96\% accuracy under professional validation. For organ labels, we constrain the VLM outputs to a predefined vocabulary of major MRI-visible organs to improve reliability and avoid unconstrained predictions.

\subsection{PET-guided Noise Modulation}
Standard diffusion bridges inject spatially uniform noise across all spatial locations~\cite{LiuG}, which is suboptimal for whole-body PET/MR translation because anatomical complexity and lesion-related variability are highly heterogeneous across regions. To address this, we introduce a PET-guided noise modulation mechanism that jointly conditions the forward corruption process on PET uptake and the region context embedding $\mathbf{c}_{\text{reg}}$.
Specifically, we use a lightweight noise modulator $\mathcal{M}_\phi$ to predict a spatial scaling map $\mathbf{S} \in [s_{\min}, s_{\max}]^{H \times W}$. The modulator extracts spatial features from PET using a two-layer convolutional encoder, incorporates anatomical context via FiLM~\cite{perez2018film} conditioning with $\mathbf{c}_{\text{reg}}$, and outputs per-pixel scaling factors. We then replace the standard uniform-noise bridge forward process
\begin{equation}
\mathbf{x}_t = \mu_t(X_0, X_1) + \Sigma_t^{1/2} \boldsymbol{\epsilon}, 
\qquad \boldsymbol{\epsilon} \sim \mathcal{N}(\mathbf{0}, \mathbf{I}),
\end{equation}
with the spatially modulated form, where $\odot$ denotes element-wise multiplication.
\begin{equation}
\mathbf{x}_t = \mu_t(X_0, X_1) + \Sigma_t^{1/2} (\mathbf{S} \odot \boldsymbol{\epsilon}),
\end{equation}
This yields a heterogeneity-adaptive forward corruption process, allowing anatomically and pathologically complex regions to receive different perturbation strengths.
\subsection{Multi-Scale PET Attention}
Because the forward process applies spatially non-uniform corruption guided by PET and anatomical context, the reverse model must recover lesion-relevant details under the same setting. To this end, we introduce PET-aware self-attention at multiple UNet resolutions. For each feature map $\mathbf{x} \in \mathbb{R}^{B \times C \times H \times W}$, the PET image is aligned by adaptive average pooling and projected by a $1 \times 1$ convolution to obtain $\mathbf{p}'$. We concatenate $\mathbf{x}$ and $\mathbf{p}'$ to form $\mathbf{z} = [\mathbf{x}; \mathbf{p}']$, and apply PET-aware self-attention:
\begin{equation}
\small
\mathbf{Q}=W_Q\tilde{\mathbf{z}},\;
\mathbf{K}=W_K\tilde{\mathbf{z}},\;
\mathbf{V}=W_V\tilde{\mathbf{z}},\qquad
\mathbf{x}_{\mathrm{out}}
=
\mathbf{x}
+
W_{\mathrm{out}}
\!\left[
\mathrm{Softmax}\!\left(\frac{\mathbf{Q}\mathbf{K}^\top}{\sqrt{d_h}}\right)\mathbf{V}
\right],
\end{equation}
where $\tilde{\mathbf{z}} \in \mathbb{R}^{B \times HW \times 2C}$ is the flattened form of $\mathbf{z}$. We use a zero-initialized $W_{\mathrm{out}}$ so that the block starts as an identity mapping and gradually incorporates PET-guided lesion cues during training.

\section{Experiments}
\begin{table}[t]
  \centering
  \caption{\textbf{Per-region quantitative results on the testset} (mean $\pm$ std). Each cell shows SSIM (\%, top) and PSNR (dB, bottom).}
  \label{tab:per-region quantitative results on the testset}
  \footnotesize
  \renewcommand{\arraystretch}{1.08}
  \setlength{\tabcolsep}{1.0pt}
  \begin{adjustbox}{max width=1.03\linewidth,center}
  \begin{tabular}{@{}l@{\hspace{3pt}}cccccc@{}}
    \toprule
    \textbf{Method}
      & \makecell[c]{\textbf{Head/Neck}\\[-2pt]\scriptsize\textit{n=\num{275}}}
      & \makecell[c]{\textbf{Thorax}\\[-2pt]\scriptsize\textit{n=\num{498}}}
      & \makecell[c]{\textbf{Abdomen}\\[-2pt]\scriptsize\textit{n=\num{617}}}
      & \makecell[c]{\textbf{Pelvis/Hips}\\[-2pt]\scriptsize\textit{n=\num{472}}}
      & \makecell[c]{\textbf{Thighs}\\[-2pt]\scriptsize\textit{n=\num{225}}}
      & \makecell[c]{\textbf{Avg}\\[-2pt]\scriptsize\textit{n=\num{2087}}} \\
    \midrule
    \multicolumn{7}{@{}l@{}}{\small\textit{General translation models}}\\[-2pt]
    Pix2Pix\cite{isola2017image}
      & \spcell{0.865}{0.063}{22.77}{1.64}
      & \spcell{0.799}{0.045}{22.31}{1.89}
      & \spcell{0.809}{0.055}{21.58}{1.75}
      & \spcell{0.835}{0.050}{22.01}{2.12}
      & \spcell{0.857}{0.037}{23.52}{1.93}
      & \spcell{0.825}{0.056}{22.22}{1.97} \\
    Palette\cite{saharia2022paletteimagetoimagediffusionmodels}
      & \spcell{0.896}{0.069}{25.83}{4.50}
      & \spcell{0.854}{0.040}{23.83}{2.86}
      & \spcell{0.856}{0.053}{23.61}{2.90}
      & \spcell{0.881}{0.051}{24.59}{3.09}
      & \spcell{0.898}{0.042}{25.44}{4.03}
      & \spcell{0.871}{0.054}{24.37}{3.41} \\
    $I^{2}SB$ \cite{LiuG}
      & \spcell{0.880}{0.062}{25.81}{1.91}
      & \spcell{0.822}{0.038}{21.67}{1.74}
      & \spcell{0.840}{0.036}{21.50}{1.31}
      & \spcell{0.855}{0.029}{23.06}{1.27}
      & \spcell{0.857}{0.027}{24.38}{2.13}
      & \spcell{0.846}{0.043}{22.77}{2.20} \\
    \midrule
    \multicolumn{7}{@{}l@{}}{\small\textit{Medical translation models}}\\[-2pt]
    PPT\cite{PPT}
      & \spcell{0.896}{0.044}{23.91}{1.44}
      & \spcell{0.811}{0.035}{21.42}{1.21}
      & \spcell{0.795}{0.041}{20.43}{1.60}
      & \spcell{0.791}{0.032}{20.93}{1.59}
      & \spcell{0.765}{0.028}{22.37}{2.09}
      & \spcell{0.808}{0.052}{21.45}{1.92} \\
    SelfRDB\cite{arslan2024selfconsistentrecursivediffusionbridge}
      & \spcell{0.849}{0.072}{22.12}{2.59}
      & \spcell{0.846}{0.071}{22.02}{2.70}
      & \spcell{0.847}{0.066}{21.95}{2.63}
      & \spcell{0.849}{0.065}{22.10}{2.65}
      & \spcell{0.857}{0.064}{22.27}{2.06}
      & \spcell{0.849}{0.068}{22.06}{2.59} \\
    ResViT\cite{Dalmaz_2022}
      & \spcell{0.910}{0.055}{26.35}{2.13}
      & \spcell{0.854}{0.038}{23.76}{2.06}
      & \spcell{0.853}{0.041}{22.89}{1.93}
      & \spcell{0.873}{0.045}{23.74}{2.42}
      & \spcell{0.860}{0.033}{24.83}{2.16}
      & \spcell{0.866}{0.047}{23.95}{2.39} \\
    \midrule
    \multicolumn{7}{@{}l@{}}{\small\textit{Our method}}\\[-2pt]
    HA-DSB(no PET)
    & \spcellsecond{0.940}{0.031}{28.59}{2.36}
    & \spcellsecond{0.890}{0.043}{24.87}{2.44}
    & \spcellsecond{0.881}{0.048}{24.06}{2.81}
    & \spcellsecond{0.931}{0.039}{27.42}{2.95}
    & \spcellsecond{0.933}{0.030}{28.51}{2.44}
    & \spcellsecond{0.908}{0.048}{26.09}{3.24} \\
    \textbf{HA-DSB(PET)}
    & \spcellbest{0.945}{0.035}{28.88}{2.48}
    & \spcellbest{0.895}{0.042}{24.92}{2.46}
    & \spcellbest{0.885}{0.048}{24.15}{2.71}
    & \spcellbest{0.932}{0.040}{27.59}{3.23}
    & \spcellbest{0.936}{0.032}{28.89}{2.49}
    & \spcellbest{0.911}{0.048}{26.25}{3.35} \\
    \bottomrule
  \end{tabular}
  \end{adjustbox}
  \vspace{-2pt}
  \begin{minipage}{0.97\textwidth}
    \scriptsize \raggedright \sloppy \setlength{\emergencystretch}{1em}
    \textsuperscript{\dag} Second-best result (SSIM and PSNR evaluated separately). 
    \end{minipage}
\end{table}
\noindent\textbf{Dataset.}
All PET/MR examinations were performed on an integrated hybrid PET/MR system (SIGNA™ PET/MR, GE HealthCare). This platform integrates a 3.0-T MRI scanner with time-of-flight PET capability. We prospectively collect whole-body imaging data from 246 subjects at an anonymous hospital. Each subject contains three co-registered modalities (LAVA, T2, and PET), all resampled to 256×256 resolution. The dataset was split into a training set of 204 subjects (20,306 slices) and a held-out testset of 42 subjects (4,162 slices). The testset comprises 21 randomly sampled subjects and an additional 21 subjects with clinically confirmed lesions, allowing us to specifically assess synthesis fidelity in pathological regions.

\noindent\textbf{Implementation Details.}
\label{sec:implementation}
For the denoising network, we employ an ADM-style U-Net~\cite{dhariwal2021adm} featuring $128$ base channels, channel multipliers of $[1, 1, 2, 2, 4, 4]$, and three residual blocks per stage. PET attention blocks are integrated at resolutions of $32 \times 32$, $16 \times 16$, and $8 \times 8$ within both the encoder and decoder. The noise modulator $\mathcal{M}_\phi$ has $s_{\min} = 0.2$ and $s_{\max} = 2.0$. The diffusion bridge uses $T = 200$ training discretization steps with $\beta_{\max} = 0.35$ and a symmetric variance schedule.
Training was conducted on two NVIDIA RTX 4090 GPUs for $50\text{k}$ iterations. We utilized the AdamW optimizer with a learning rate of $10^{-4}$, a total batch size of $16$, and an EMA decay of $0.9998$. 

\begin{figure}[!t]
  \centering
  \includegraphics[width=1\linewidth]{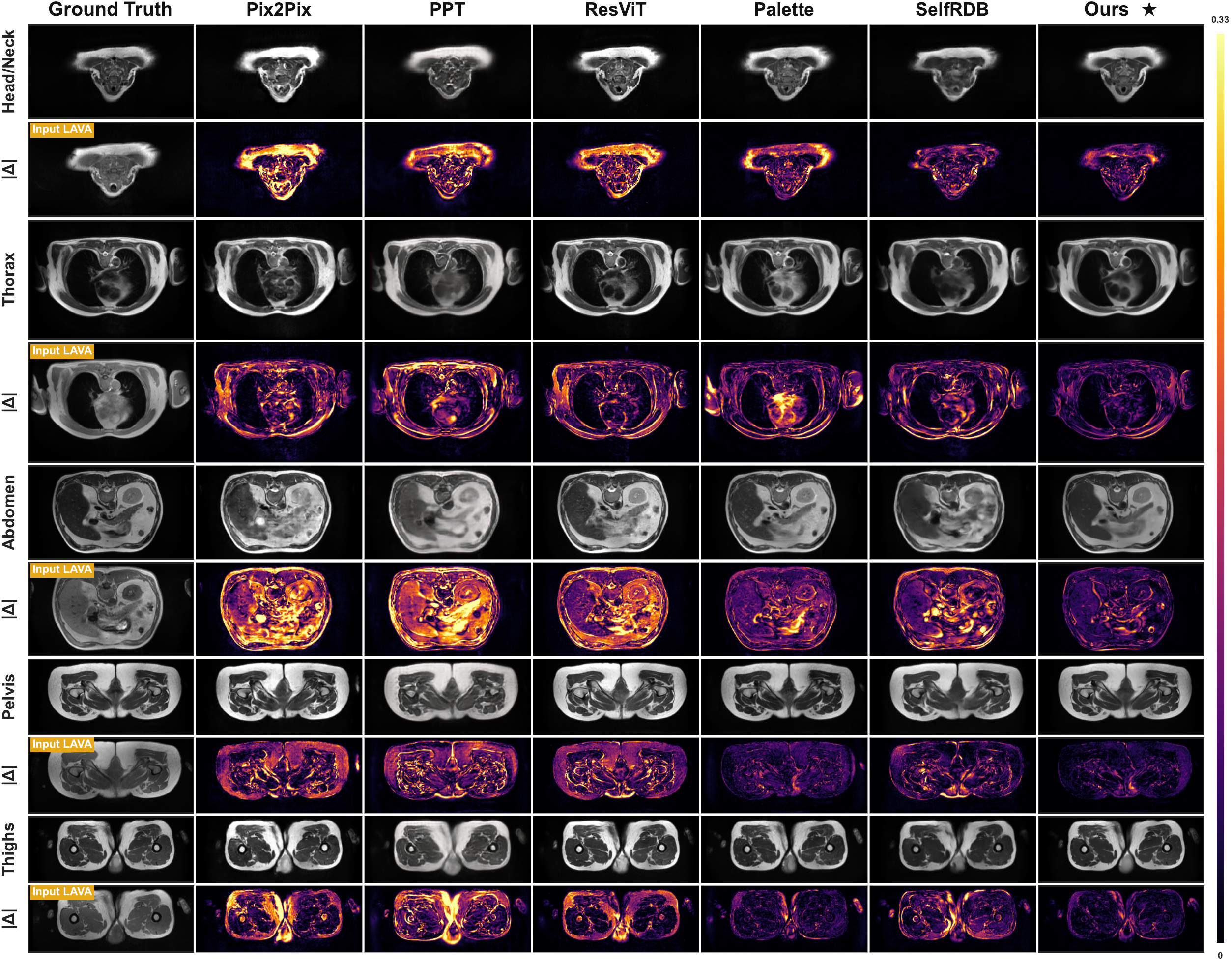}
  \caption{\textbf{Qualitative comparison across five anatomical regions.}
           Top row: translated T2w results; bottom row: absolute error maps
           $|\hat{\mathbf{x}}-\mathbf{x}_{\mathrm{GT}}|$.}
  \label{fig:qualitative_main}
\end{figure}

\begin{figure}[t]
    \centering
    \includegraphics[width=1\linewidth]{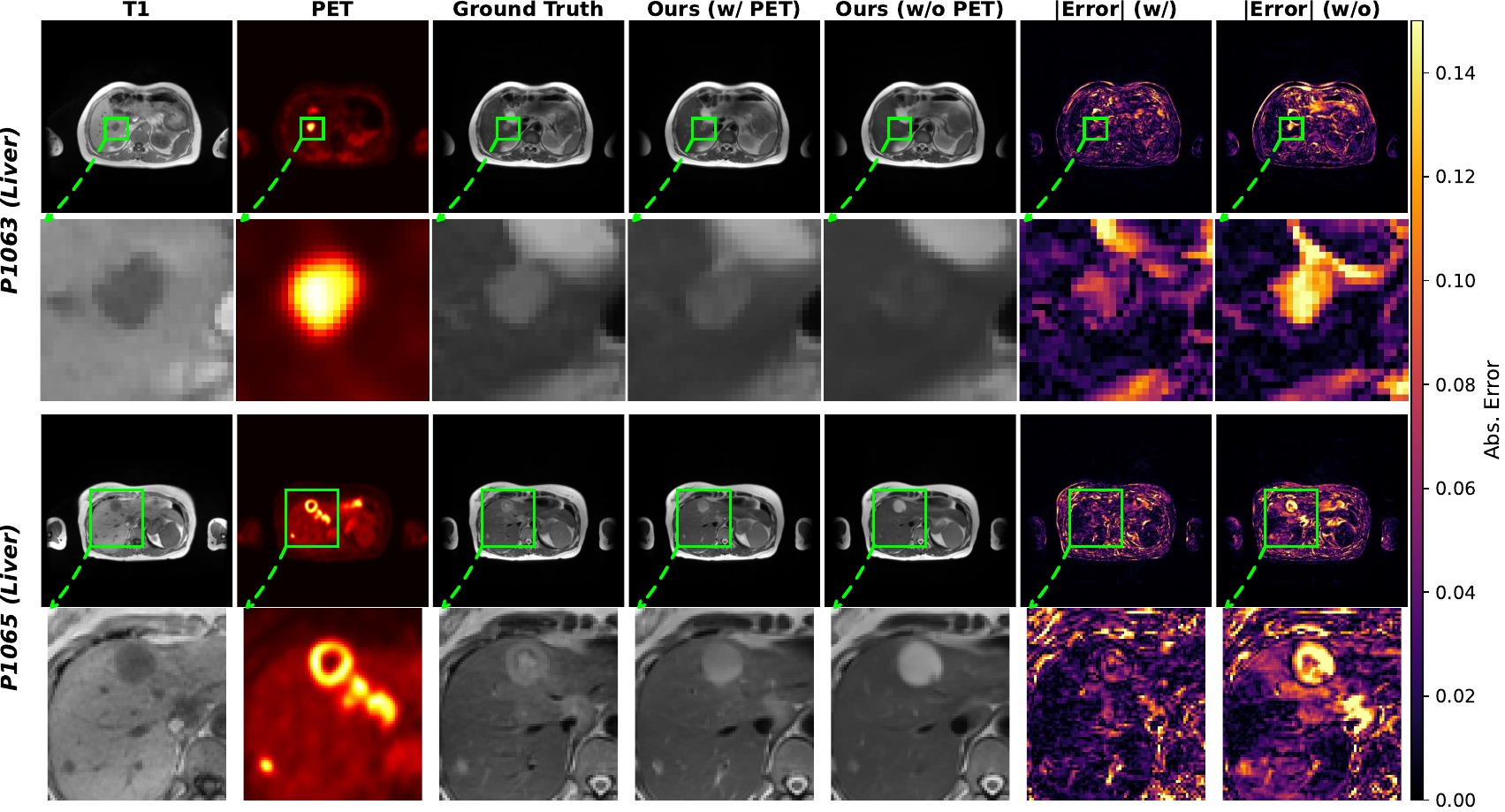}
    \caption{\textbf{Effect of PET guidance on pathological cases (liver lesions).}
           From left to right: T1, PET, GT, ours (w/ PET), ours (w/o PET),
           and their absolute error maps; zoomed views correspond to the
           green boxes (shared color scale).}
    \label{fig:pet_effect_diagram}
\end{figure}

\noindent\textbf{Comparison with SOTA methods.} 
\noindent We compare HA-DSB against five representative baselines from GAN-based, diffusion-based, and bridge paradigms, all trained on whole-body data. As shown in Table~\ref{tab:per-region quantitative results on the testset}, HA-DSB achieves the best average SSIM and PSNR, outperforming the strongest baseline, ResViT. Compared with the basic diffusion bridge baseline I$^2$SB, HA-DSB achieves consistent improvements across all anatomical regions, clearly demonstrating the effectiveness of the proposed region-aware conditioning mechanism. Simpler regions, such as Head/Neck, Pelvis/Hips, and Thighs, achieve SSIM above 93\%, while structurally complex regions such as the Abdomen also show clear gains. These results suggest that region guidance helps the bridge learn region-dependent transport paths. Qualitative comparisons in Fig.~\ref{fig:qualitative_main} further support this observation.

\begin{table}[b]
  \centering
  \caption{Effect of PET guidance on pathological cases. $\Delta$ denotes
           improvement from PET guidance. Best in \textbf{bold}.}
  \label{tab:ablation_sick}
  \scriptsize
  \renewcommand{\arraystretch}{0.95}
  \setlength{\tabcolsep}{2.6pt}
  \begin{tabular}{lcc}
    \toprule
    \textbf{Variant} & \textbf{PSNR (dB)} & \textbf{SSIM (\%)} \\
    \midrule
    HA-DSB(no pet) & \pmNum{23.46}{2.25} & \pmNum{86.4}{3.9} \\
    HA-DSB(pet) & {\bfseries\pmNum{24.26}{2.67}} & {\bfseries\pmNum{88.1}{3.9}} \\
    \midrule
    $\Delta_{\text{PET}}$ & \textbf{+0.80 (vs +0.16)} & \textbf{+1.7 (vs +0.3)} \\
    \bottomrule
  \end{tabular}
\end{table}
\noindent\textbf{Effect of PET guidance.}PET guidance (HA-DSB(pet) vs.\ HA-DSB(no pet)) brings only modest gains in overall SSIM and PSNR on the testset (21 patients, 2{,}087 slices). This is likely because lesion regions occupy only a small fraction of the total volume, and their improvements are diluted by the dominant healthy tissue where PET provides limited additional information.
To better isolate the impact of PET guidance on pathological regions, we further evaluate performance on a separate lesion-confirmed cohort of 21 patients. A radiologist identified slices covering disease-bearing organs, resulting in 526 lesion-containing slices out of 2{,}075 (Table~\ref{tab:ablation_sick}). 
On this pathological subset, PET guidance yields markedly larger improvements in both PSNR and SSIM than on the testset, indicating that the noise modulation and multi-scale attention primarily enhance lesion-related regions. Fig.~\ref{fig:pet_effect_diagram} further illustrates two representative cases.

\section{Conclusion}
\noindent In this work, we proposed HA-DSB for PET-guided whole-body MR translation. By combining region context embeddings, PET-guided noise modulation, and multi-scale PET-aware attention, HA-DSB addresses anatomical and pathological heterogeneity in whole-body PET/MR. Experiments show improved translation quality across body regions, with additional gains in lesion-containing areas. These results indicate that heterogeneity-adaptive conditioning is effective for modeling region-dependent structural variability and recovering lesion-relevant details in challenging whole-body settings. More broadly, HA-DSB provides a practical bridge-based framework for robust whole-body medical image translation.
\subsubsection*{\textbf{Disclosure of Interests}}
\footnotesize
The authors have no competing interests to declare that are relevant to the content of this article.

%
%
%
%

\end{document}